# Phorecaster365: A Human-Supervised Reference Architecture for Hybrid Pharmaceutical Sales Forecasting and Planning Decision Support

Houman Kazemzadeh[1,2] Kamyar Naderi[2]

[1] Izlek Technology, Ankara, Türkiye

[2] Xylemed, Dubai, United Arab Emirates

Correspondence: Houman Kazemzadeh | houman@xylemed.com

**Document note.** This technical report describes a reference architecture and synthetic development experience. Forecasts support accountable planning review. Autonomous purchasing, manufacturing release, medicine allocation, and patient-level clinical decisions are outside its intended use.

## Abstract

Pharmaceutical sales forecasts inform planning across products, regions, and distribution channels, yet their interpretation depends on inventory availability, transaction semantics, product lifecycle, and the information available when each forecast is issued. A model prediction alone does not preserve these conditions or establish whether a forecast is suitable for operational use. We present Phorecaster365, a human-supervised reference architecture that connects enterprise resource planning data to reviewable pharmaceutical sales forecasts. The architecture separates source ingestion, product–region series construction, temporally valid feature generation, statistical and machine-learning modeling, ensemble formation, uncertainty assessment, planner review, and lifecycle governance. Its central intermediate representation is a forecast context package that preserves the source snapshot, forecast target, temporal cutoff, available covariates, data-quality state, and hierarchy version. A corresponding forecast evidence package links predictions to model and calibration versions, exceptions, human adjustments, and publication history. Development experience with a synthetic panel of 10,950 daily records across 30 product–region series informs the design. Historical experiment summaries are retained only as descriptive evidence of development because their evaluation does not establish independent predictive validity. We specify a rolling-origin evaluation protocol, baseline and ablation requirements, uncertainty and robustness assessments, and a staged pathway from synthetic testing to use in governed planning. The contribution is an implementation-neutral system design and validation framework; the report does not establish real-world forecasting accuracy, comparative superiority, or operational benefit.

**Keywords:** pharmaceutical sales forecasting; hybrid forecasting; ensemble learning; enterprise resource planning; time series; uncertainty quantification; human oversight; forecast provenance; model governance

## 1 Introduction

Pharmaceutical planning requires estimates of future sales at a grain that can be connected to products, locations, and planning horizons. Recorded transactions are shaped by commercial activity and by the supply system through which medicines reach customers. A decline in sales may reflect lower demand, unavailable stock, delayed fulfillment, a change in distribution coverage, or a revision to the reporting process. These explanations have different planning implications, even when they produce similar time-series patterns.

The operational task, therefore, extends beyond fitting a predictive model. Planners need to know what quantity is being forecast, which events were available at issuance, whether the recent history is complete, how uncertainty changes over the horizon, and whether a forecast has been reviewed. They also need a reproducible connection between a published number and the source records, model configuration, and human decisions that produced it. Without this connection, later evaluation may compare forecasts against a different version of history or attribute a manual adjustment to the model.

We propose Phorecaster365 as an intermediate layer between enterprise resource planning (ERP) systems and planning workflows. Its initial modeling scope is daily product–region sales forecasting over a 30-day horizon. The architecture is modular: a statistical seasonal signal and tree-based predictors can be evaluated within a common temporal framework, while ingestion, uncertainty, review, and publication retain explicit interfaces. A forecasting component can be replaced without redefining the target's meaning or removing the controls around its use.

Phorecaster365 is designed around the operational requirements of pharmaceutical forecasting. Each forecast is linked to the data and information available at the time of issuance, together with its target definition, feature configuration, model version, calibration method, and review history. Temporal provenance and versioned forecast packages distinguish model-generated predictions from planner-approved forecasts and preserve the traceability of subsequent adjustments. This structure supports reproducible evaluation, accountable human oversight, and controlled use within pharmaceutical planning workflows.

This report makes four architectural contributions. First, it specifies a temporal data contract that distinguishes observed sales, availability constraints, and information known at the forecast cutoff. Second, it defines context and evidence packages that connect model execution to planner review and downstream publication. Third, it describes an evaluation design that jointly addresses multi-step forecasting, ensemble selection, uncertainty, and hierarchy consistency. Fourth, it provides a deployment and governance framework with explicit evidence requirements at each stage. These contributions concern system integration and testable operating requirements; no new forecasting algorithm is claimed.

## 2 Related work and design rationale

Forecasting methods offer complementary representations of temporal structure. Prophet models trend, seasonal components, and holiday effects in a framework intended to support forecasting at scale. XGBoost provides a scalable gradient-boosted tree system. Random forests combine randomized trees, while gradient boosting constructs an additive predictor through successive fitting steps. These model families provide plausible components for pharmaceutical sales forecasting when their inputs and evaluation match the operational task.

The forecasting literature also makes clear that performance depends on how errors are measured and aggregated. Scale-dependent errors, percentage errors, and scaled errors answer different questions, particularly when series differ in volume or contain zeros. Forecast reconciliation addresses another requirement: predictions across a product and geographic hierarchy should respect the aggregation relationships used by the business. Time-series uncertainty methods, including ensemble-based conformal approaches, address interval construction under specified assumptions about dependence and error behavior. None of these methods removes the need to test the complete deployed workflow.

Phorecaster365 uses these established foundations to address a systems problem. An operational forecast must preserve the meaning of transactions and the availability of historical information, expose the reliability of its inputs, and distinguish a model recommendation from an authorized plan. These requirements remain relevant regardless of whether the prediction engine uses statistical models, trees, or a later model family. The design consequently places model comparison inside an architecture with stable data, evidence, and publication contracts.

## 3 Intended use and system requirements

The intended users are demand planners, commercial analysts, supply-planning teams, and authorized reviewers working with aggregated pharmaceutical sales data. Phorecaster365 prepares forecasts and supporting evidence for planning review. It may help organize exceptions and compare scenarios, but the resulting forecast does not itself determine a purchase quantity, production schedule, or distribution allocation. Such decisions require additional objectives and constraints, including lead times, available inventory, expiry dates, service policies, and capacity.

The initial target is the number of units sold under a declared transaction convention. Forecasting revenue requires a separate price and currency convention. Forecasting unconstrained demand requires explicit treatment of lost sales, backorders, or another source of demand information. A system configuration must name its target before ingestion, modeling, and evaluation are specified; these quantities are not interchangeable.

These conditions are normative properties of the reference architecture. Their inclusion does not imply that all controls were implemented or tested in the synthetic prototype. Deployment claims require evidence that the corresponding condition has been satisfied in the target environment.

**Table 1.** Requirements and observable acceptance conditions.

| Requirement | Architectural treatment | Acceptance condition |
|---|---|---|
| **Temporal validity** | Store event time and knowledge time; compute features at historical cutoffs | Every input can be reconstructed from information available at issuance |
| **Target consistency** | Version units, returns, cancellations, and aggregation rules | Training, evaluation, and publication use the same measure |
| **Traceable output** | Link context, models, calibration, review, and export | A published forecast resolves to its complete evidence package |
| **Visible uncertainty** | Record interval method, level, horizon, and calibration status | Unavailable or invalid intervals are distinguishable from valid estimates |
| **Human accountability** | Preserve original values and signed adjustments | Every published revision has an authorized review history |
| **Controlled failure** | Hold, quarantine, or issue an approved fallback | Missing or malformed forecasts cannot silently enter planning |

## 4 Data semantics and the forecast context package

### 4.1 Canonical entities and temporal provenance

The ingestion boundary receives sales events, reversals and returns, product and location masters, inventory snapshots, and commercial calendars. Source records remain associated with stable identifiers and a source snapshot. The canonical model separates the time an event occurred from the time it became available to the forecasting system. A corrected invoice posted after a historical forecast cutoff may relate to an earlier sales date, but it cannot be treated as if the correction were already known at that cutoff.

This distinction also applies to master data and commercial plans. A region mapping needs an effective date and a version. A promotion requires both its planned execution dates and the time its plan became available. A forecast replay reconstructs the earlier information state rather than joining every historical transaction to the latest product master or the final realized promotion calendar. Late-arriving actuals may be used to score a past forecast under a declared outcome-finalization policy, while its original input snapshot remains unchanged.

Data validation checks unique keys, date ranges, unit conversions, duplicate events, missing feeds, reversals, and consistency between transaction and master records. Critical failures hold the affected scope. A localized failure may quarantine a series while leaving independently valid series available for processing. Any portfolio total must disclose incomplete coverage; it cannot imply completeness after excluding failed segments.

### 4.2 Sales availability and demand censoring

Observed sales measure realized transactions. When stock is unavailable, realized sales can be smaller than customer demand. An inventory field alone does not identify how many sales were lost. Phorecaster365 therefore records availability and censoring status without automatically converting them into a point estimate of unconstrained demand. If a separate demand-reconstruction model is introduced, its assumptions, outputs, and uncertainty require independent evaluation.

Zeros and missing values receive distinct meanings. An observed zero on an open, fully reporting day can be a valid sales observation. A closed outlet, an incomplete feed, a stockout, and an unresolved reversal each require a different status label. Imputation preserves the original state and its method; it does not silently replace missing observations with measured values. Returns remain distinct from gross sales, and a forecast of net units may legitimately have different support from a forecast of nonnegative gross sales.

Pack sizes, dosage forms, and strengths require controlled product identifiers. Aggregating different medicines into a portfolio unit count is a commercial reporting convention and does not establish therapeutic equivalence. Monetary totals require explicit assumptions about currency and price. Product substitutions, launches, and discontinuations require effective dates and reviewable business context; they cannot be reliably inferred from a product name alone.

### 4.3 Context representation

We define the forecast context package as the stable input representation shared by the forecasting components. It contains the target specification, historical series, observation states, permitted covariates, forecast horizon, and temporal and hierarchy metadata. It also records the data-quality decision and the scope of any exclusions. This representation allows changing a model while retaining the same task definition and information boundary.

**Table 2.** Proposed forecast context package.

| Content group | Required information |
|---|---|
| **Task identity** | Run identifier, organization scope, product and region identifiers, target measure, unit, daily grain, and horizon |
| **Temporal state** | Issuance cutoff, source snapshot, ingestion time, outcome-finalization policy, and applicable calendar |
| **Historical observations** | Values with observed, zero, censored, imputed, corrected, missing, or unavailable status |
| **Product and geography** | Pack and unit mappings, lifecycle state, region mapping, and hierarchy version |
| **Covariate availability** | Historical values, approved future values, knowledge timestamps, and scenario identifiers |
| **Quality and lineage** | Validation findings, excluded scope, source references, transformation version, and approval state |

Scenario inputs remain separate from the operational information set. For example, a proposed promotion can be explored in a scenario package, while the baseline package contains only the approved plan known at issuance. A difference between scenario forecasts is a model-conditioned comparison. It is not an identified causal effect of the promotion.

## 5 Reference architecture

The proposed architecture contains a data path and a governance path, as illustrated in Figure 1. The data path transforms validated enterprise records into forecast evidence for review. The governance path controls access, versioning, exceptions, model changes, and publication. Their separation allows a model service to produce a valid computational output without giving it authority to publish a planning decision.

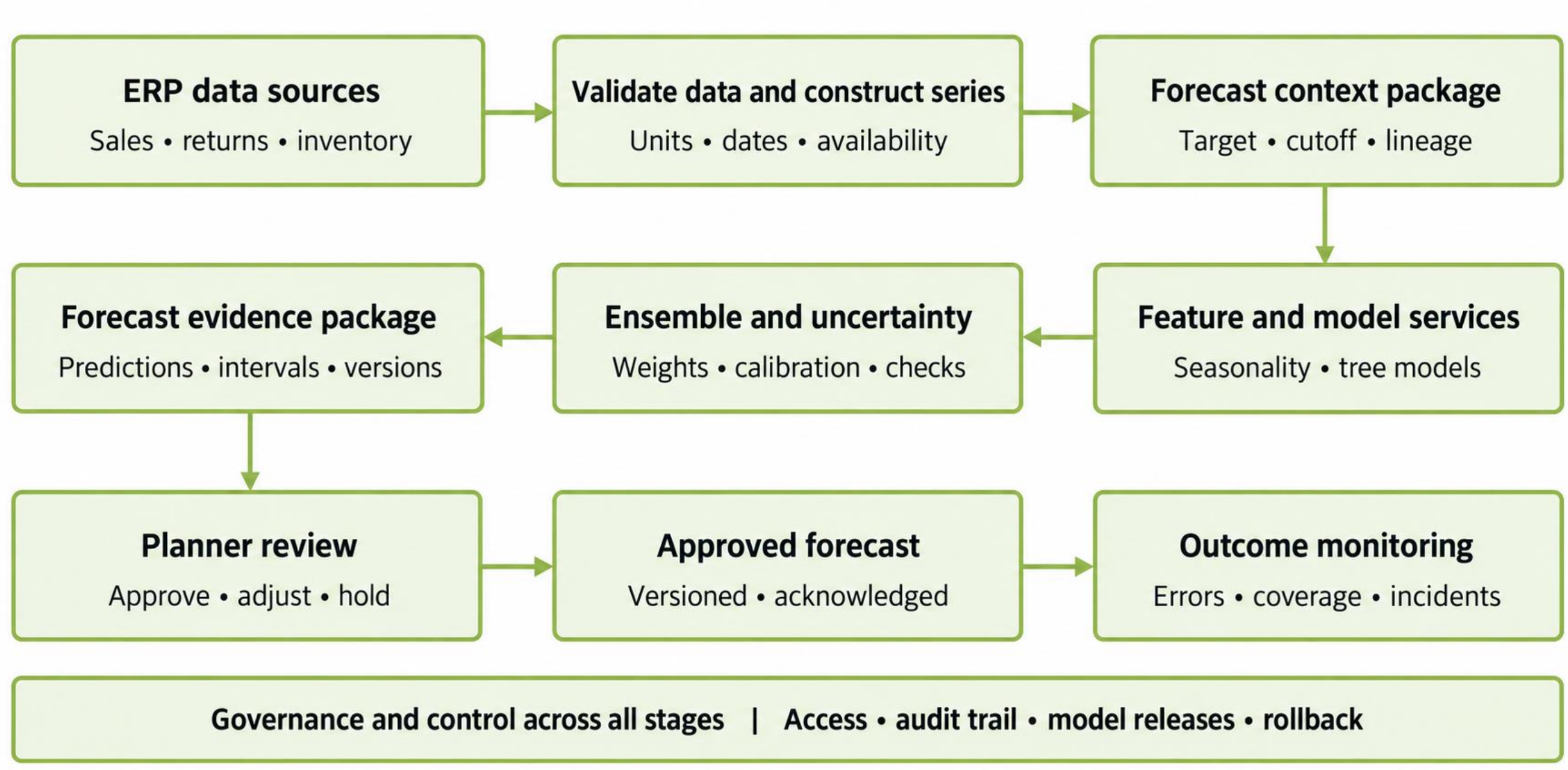

**Figure 1.** Phorecaster365 reference architecture. Solid arrows show the forecast production path. The governance layer applies to every stage; outcome monitoring can initiate candidate evaluation, but does not authorize a model release. The diagram describes the target architecture.

### 5.1 Ingestion and series construction

A scheduled ingestion process freezes the source snapshot and runs the data contract. The series builder then aggregates valid events into daily product–region observations, attaches observation states, and constructs the declared calendar. Gross sales, returns, and net sales remain separate measures. Source corrections create a new data version rather than rewriting the evidence associated with an earlier forecast.

The service preserves launches, retirement dates, region changes, and product remapping. A newly introduced series may require a pooled or analog-based strategy, while a discontinued product requires a declared run-out policy. A perfectly regular date index is a computational representation; it does not by itself establish that all dates contain observed sales.

### 5.2 Feature and modeling services

The feature service materializes only inputs permitted by the context package. The modeling service consumes this representation and returns candidate predictions, fitting metadata, and failure status through a common interface. Temporal validation controls the feature definition, model configuration, and ensemble weighting together. A model that cannot produce a valid forecast for the requested horizon returns a visible failure or fallback request rather than a partial array without explanation.

The reference architecture permits local models for individual series and pooled models across related series. The development experience described in Section 9 concerns product–region modeling; pooled learning remains an extension requiring a separate comparison. Organization boundaries must be preserved in any multi-organization deployment, including training, feature storage, evaluation, and retrieval of forecast history.

### 5.3 Uncertainty and consistency services

The uncertainty service receives predictions and calibration evidence with explicit versions. It records whether intervals are available, insufficiently supported, stale, or withheld. A consistency service checks horizon completeness, dates, units, finite values, and hierarchy relationships. A restriction to nonnegative predictions applies only when it is appropriate for the declared target. Any transformation, reconciliation, or clipping that changes a forecast must be included in the evaluated pipeline.

### 5.4 Review and publication services

The review service exposes predictions with context, limitations, and exceptions. The publication service accepts only versions that satisfy the configured approval rule. Export uses an idempotent identifier and records downstream acknowledgment. Failed exports remain distinguishable from forecasts that were approved and successfully received. A retry must not create duplicate planning records.

The design permits batch integration with an existing planning system or a separate review application. It does not depend on a specific ERP vendor, database product, dashboard framework, or hosting platform. Those choices require site-specific integration testing and do not constitute demonstrated capabilities of the present prototype.

## 6 Hybrid forecasting methodology

### 6.1 Statistical signals and tree predictors

The development trajectory combined a Prophet seasonal forecast with XGBoost, random forest, and gradient-boosting predictors. In the later hybrid configuration, the statistical forecast serves as an input signal to the tree predictors. The final weighted combination is formed from the tree-model outputs. This distinction matters: adding a Prophet-derived feature is different from assigning Prophet a separate ensemble weight. A Prophet-only forecast remains an essential comparator in the proposed benchmark.

The statistical component can represent recurring calendar patterns and changes in trend. Tree models can use lagged sales, historical rolling summaries, calendar variables, and admissible commercial context to represent nonlinear relationships. Their errors can still be strongly correlated because they share data and features. Combining model families, therefore, creates a testable opportunity for improvement; it does not guarantee diversification or better generalization.

Annual seasonality is particularly difficult to substantiate with a single year of data. When earlier training windows contain only part of that year, fitting a flexible annual component may absorb transient structure. Future evaluation should test whether the component adds value across multiple seasonal cycles and whether a simpler weekly or calendar representation is sufficient.

### 6.2 Temporal feature policy

For a forecast issued after the last available day, observed target lags and rolling summaries may use only values that were available by that issuance cutoff. In training, a feature associated with a target day must exclude that day's target. This policy applies to derived seasonal signals as well as direct target statistics. A Prophet prediction used as a training feature should be generated without fitting on the target it is intended to predict, using temporal out-of-fold forecasts that match the intended prediction task.

Future calendar indicators can be known in advance, but future realized sales, replenishments, inventory levels, and commercial outcomes generally are not. Approved plans may enter as known-future inputs only with documented knowledge times. Replacing a plan with its eventual realized value changes the forecasting problem and invalidates the replay as a simulation of issuance-time performance.

Transformations also respect this boundary. Imputation rules, feature selection, scaling (where used), outlier handling, and target transformations are fitted within the appropriate temporal training window. A preprocessing step learned on the full series can leak information even when the final predictor uses an apparently chronological split.

### 6.3 Ensemble formation and multi-step prediction

The candidate ensemble combines member forecasts using nonnegative weights that sum to one. Equal weighting provides a transparent reference. Performance-based weighting is estimated from earlier validation predictions using a predeclared loss and pooling rule. A small validation set can make detailed series-specific or horizon-specific weights unstable, so shrinkage toward equal weights or a pooled strategy should be evaluated. No weighting scheme is selected using the outer test outcomes.

The prototype produced 30-day recursive forecasts. During recursive issuance, future target lags and rolling summaries are updated using earlier predictions when actual observations are unavailable. Evaluation must preserve this behavior for the full horizon. Feeding observed test-period sales back into later steps would instead evaluate a sequence of updated forecasts and could not be presented as a single forecast issued for the entire 30-day period.

A direct strategy, with horizon-specific prediction targets, is a planned comparator. It may avoid some recursive error propagation, but requires enough training examples at each horizon. The architecture explicitly records the forecast strategy, so that a direct, recursive, or periodically refreshed forecast cannot be silently substituted during evaluation.

### 6.4 Hierarchy consistency and fallback behavior

Product–region forecasts may be aggregated to product totals, region totals, and portfolio summaries. Bottom-up aggregation is a straightforward reference; reconciliation methods can be compared where forecasts are also generated at higher levels. Any learned reconciliation parameters are estimated inside the training boundary. Forecasts must remain interpretable under the declared unit and hierarchy conventions.

Point coherence does not imply probabilistic coherence. Summing the lower and upper endpoints of the marginal intervals does not generally yield a calibrated interval for the total. Aggregate uncertainty requires a method that captures cross-series dependence or a separate evaluation at the aggregate level. Similarly, summing daily point forecasts produces a cumulative point estimate, but a lead-time interval requires joint path information or a separately evaluated cumulative-target procedure.

Fallback policies are explicit and evaluated. Seasonal-naïve forecasts can support series with adequate weekly history; sparse or newly launched products may need an approved pooled or analog estimate. If neither the main model nor the fallback is supported, the service holds the forecast for review. Reusing the last approved plan is a business continuity action with an age limit, not a new model forecast.

## 7 Forecast evidence and planner interaction

### 7.1 Evidence package and lineage

The forecast evidence package is the durable output of a forecasting run. It links the context package to the actual predictions, ensemble configuration, calibration state, model versions, quality checks, and review decisions. Figure 2 illustrates the dependency structure. Its purpose is to make a published value reconstructible and to preserve the distinction between measured inputs, modeled outputs, scenario assumptions, and human adjustments.

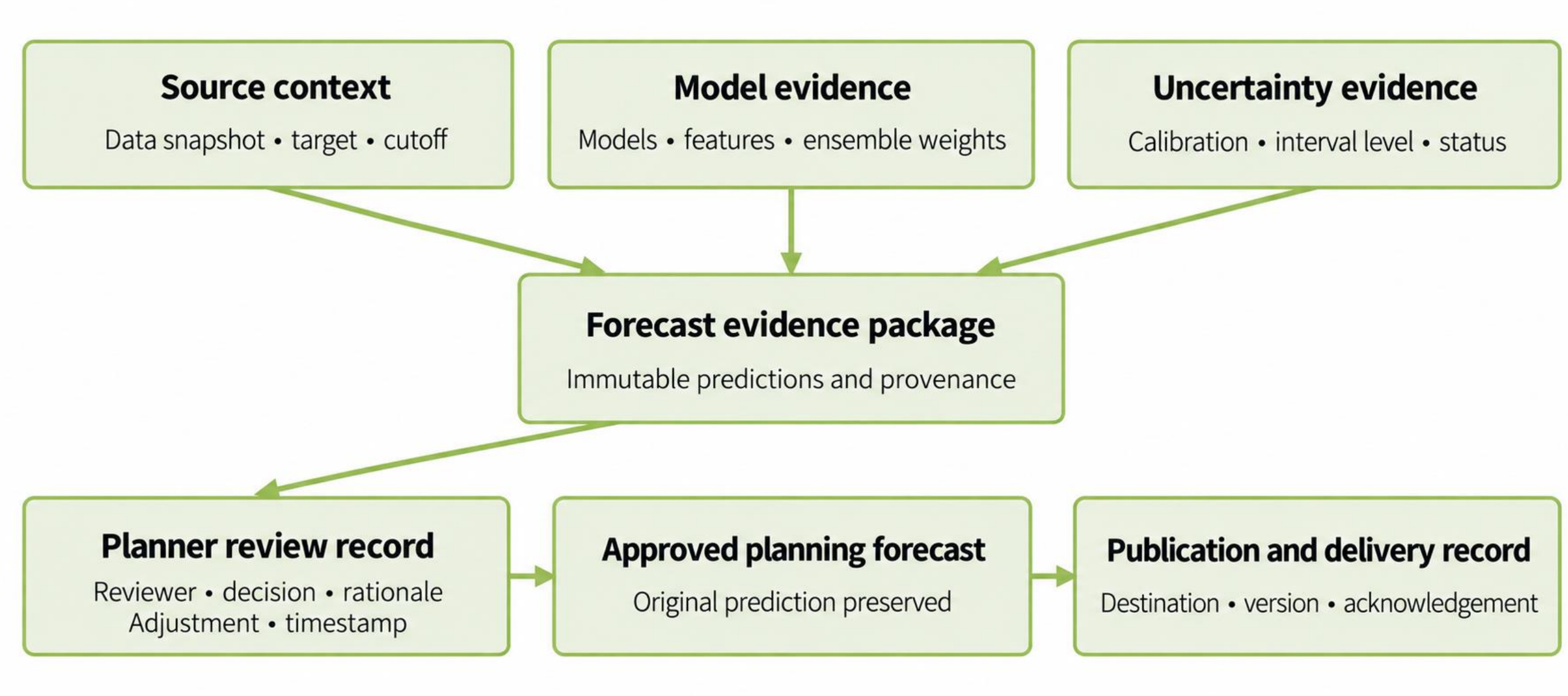

**Figure 2.** Evidence relationships for a published forecast. A forecast package points to its input and model evidence; review creates a versioned planning decision, and publication records the approved version and downstream acknowledgment. Original model predictions remain available after adjustment.

An interval entry includes its nominal level, target, horizon, method, and calibration period. A missing interval has a status rather than fabricated endpoints. A forecast entry also identifies whether it is a baseline, scenario, fallback, or human-adjusted plan. These distinctions remain available in machine-readable exports and in the review interface.

### 7.2 Review workflow

The planner workspace prioritizes exceptions based on declared business criteria, such as affected volume, recent bias, forecast changes, interval width, availability concerns, or upcoming product lifecycle changes. The queue must display why an item was prioritized. Materiality and urgency should be configurable; a high-volume product is not automatically the most important medicine for service continuity.

A reviewer can inspect recent history, forecast paths, data status, known events, prior approved forecasts, and performance at relevant horizons. The reviewer may accept, adjust, defer, or escalate. Adjustments record the replacement value, reason, supporting evidence, author, and timestamp. An adjustment creates a new version and retains the original model forecast. Approval and publication are separate events, so a forecast can be reviewed without being mistaken for a successfully delivered plan.

Model explanations describe associations in the fitted predictor. A holiday component or feature attribution may help a planner understand model behavior, but it does not prove why sales changed. Scenario explanations must name the changed assumptions. Natural-language explanations, if added later, should be generated from the evidence package and evaluated separately; a language model is not required by the present forecasting architecture.

### 7.3 Measuring human adjustments

Forecast value added compares the loss of an earlier forecasting stage with the loss after a subsequent adjustment on matched series, origins, horizons, and actuals. Under a convention where positive value denotes improvement, it is the earlier loss minus the later loss. The evaluation preserves the baseline, machine, adjusted, and published forecasts rather than retaining only the final number.

Reviewer-level comparisons require care because difficult cases may be assigned selectively. An association between overrides and errors is not sufficient to determine whether the reviewer caused the change. Reporting by reason, difficulty, horizon, and selection policy is more informative than a simple reviewer ranking. A prospective pilot should also measure review time, unresolved exceptions, and instances where users appropriately rejected an unsupported forecast.

## 8 Evaluation protocol

### 8.1 Study questions and temporal design

The evaluation should answer three distinct questions: whether the forecast adds predictive value at issuance, whether its uncertainty estimates are useful, and whether the complete workflow supports reliable planning. Predictive accuracy alone cannot establish workflow benefit, and a technically reliable deployment can still deliver poor forecasts.

Figure 3 specifies the temporal separation required for a confirmatory replay. An outer origin represents a forecast that could have been issued at a historical cutoff. Model selection and ensemble weighting use earlier inner folds. Calibration uses prediction errors not used to fit or select the point predictor. The following test horizon remains unavailable until the forecast is frozen. The protocol, targets, candidate families, metrics, and aggregation rules are fixed before the confirmatory assessment.

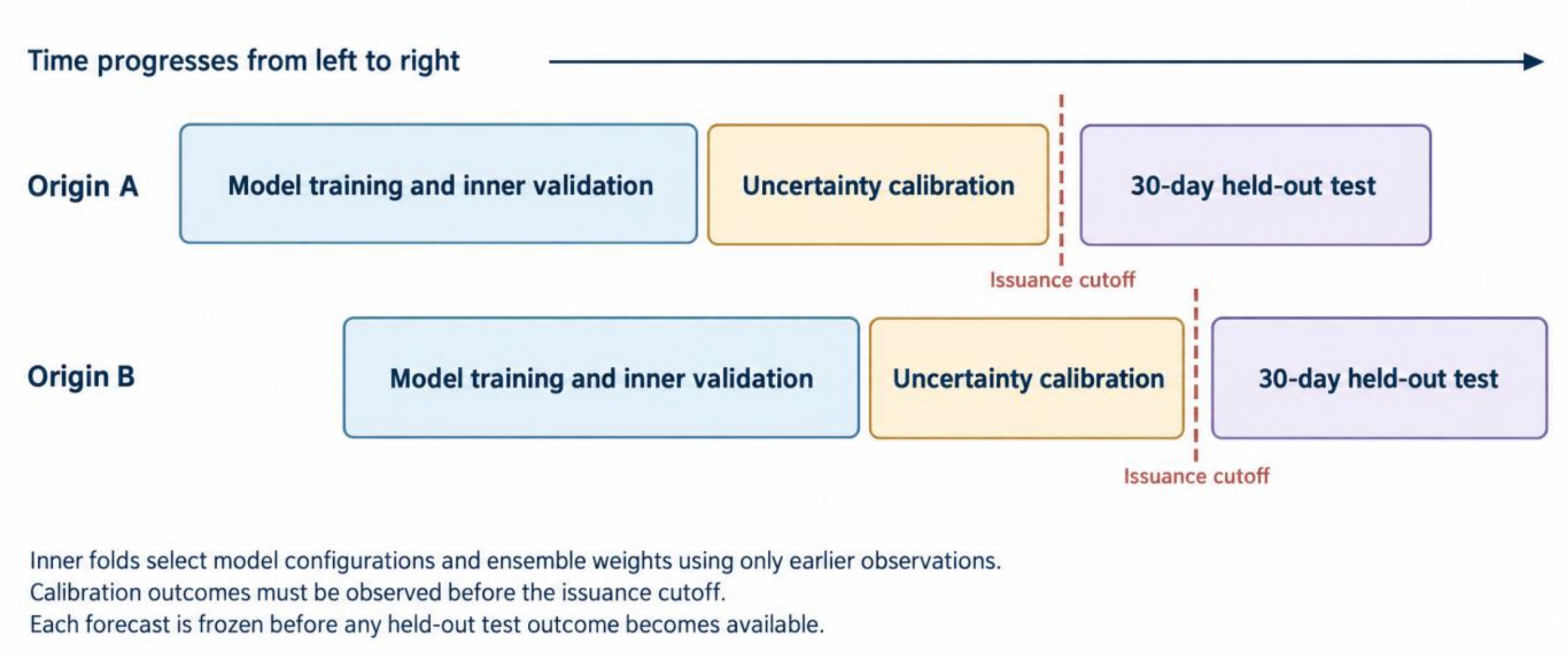

**Figure 3.** Proposed temporal evaluation design for a fixed 30-day forecast. Inner selection precedes calibration, and the outer test follows the issuance cutoff. Calibration errors must come from genuinely issued predictions whose outcomes have matured before that cutoff. Each outer origin repeats the declared procedure; the diagram is schematic and does not prescribe block lengths.

There are two defensible calibration designs with different costs. A strict split design freezes the point model before a separate calibration block and keeps it fixed through the outer test; this preserves a clear separation but sacrifices recent data for model fitting. An operational replay design generates calibration residuals from earlier forecast origins under the same declared refitting policy. It must ensure that every residual was produced without access to its outcome and that the complete outcome horizon was available before interval issuance. These designs should not be described as statistically interchangeable.

Refitting a selected model on the calibration labels and then retaining its earlier split-calibration residuals changes the fitted predictor. Validity cannot be assumed to survive that change. Whichever design is used, temporal dependence, short histories, and regime changes require empirical coverage assessment. A nominal interval level is not a distribution-free coverage guarantee for an arbitrary sales process.

## 8.2 Baselines and ablations

All methods are compared at identical origins and horizons using the same historical information boundary. The baseline suite includes last-value prediction, seven-day seasonal naïve, a declared moving-average forecast, Prophet alone, the individual tree models, an equal-weight ensemble, the performance-weighted candidate, and the incumbent planning forecast when a historical version exists. Missing forecasts and fallback use are retained in reporting.

**Table 3.** Planned ablations and the question each isolates.

| Comparison | Question |
|---|---|
| **Hybrid model versus the same model without the Prophet signal** | Does the statistical feature add incremental predictive value? |
| **Full feature set versus removal of lagged summaries** | How much does performance depend on historical target structure? |
| **Calendar model with and without holiday information** | Does the approved holiday calendar improve the defined horizon? |
| **Equal versus validation-based weights** | Does learned weighting improve on a simpler combination? |
| **Recursive versus direct forecasts** | Does forecast strategy change horizon-dependent error and reliability? |
| **Unreconciled versus reconciled outputs** | Does hierarchy consistency improve useful accuracy without unacceptable segment regressions? |
| **Model forecast versus reviewed forecast** | Do planner adjustments add value under the observed review policy? |

Removing a feature while changing tuning, training data, or forecast strategy does not isolate that feature's contribution. Ablations therefore preserve the remaining evaluation conditions. The retained development summaries do not provide a controlled ablation study; this table defines prospective comparisons.

### 8.3 Metrics and aggregation

MAE and RMSE report errors in target units. WAPE describes total absolute error relative to total absolute actual volume over a declared evaluation set. MASE scales test error by a training-period naïve error; for the weekly variant, the denominator uses seven-day differences computed only from the training history. MAPE is unsuitable when actuals are zero or near zero. sMAPE also requires an explicit convention for cases where both the actual and the prediction are zero. Undefined denominators are reported with their affected scope, rather than being silently replaced with an arbitrary constant.

Bias uses the convention of prediction minus actual, so positive values indicate overforecasting. The study reports bias by series and horizon because opposite regional errors can cancel in a portfolio mean. Both macro summaries and exposure-oriented summaries are needed: macro averages describe a typical series, while pooled WAPE reflects total absolute error relative to total volume. An average of series-level WAPE values is not the same statistic as pooled WAPE.

Uncertainty evaluation reports empirical coverage, interval width, and an interval score that penalizes both excessive width and misses. Results are stratified by horizon, product, region, availability state, and demand regime, where sample sizes permit. Publication success, missing-output rate, fallback rate, data freshness, runtime, and acknowledgment latency describe system reliability. Forecast errors from successful runs alone cannot characterize the service planners experience.

### 8.4 Dependence and confirmation

Rolling windows may overlap, and series may share product, calendar, or regional shocks. Window-level losses are therefore not independent replicates. Uncertainty in comparative metrics should be estimated using dependence-aware resampling or an inferential design matched to the data. Origin-level blocks can preserve common shocks across series; the choice and block length need justification. The 420 retained window summaries described below must not be treated as 420 independent experimental units.

A final, untouched period or a later data vintage provides confirmation once design choices are frozen. If outcomes from that period lead to a redesign, subsequent performance requires a new confirmation set. Benchmark superiority is reported only when supported by matched comparisons and practically meaningful effect sizes. Operational gains require a separate evaluation of the downstream planning process.

## 9 Synthetic development experience

### 9.1 Dataset characterization

The synthetic panel contains 10,950 daily records covering 1 January through 31 December 2023. It combines six products with five named regions. Every product–region pair contains 365 observations, giving 30 complete series. These labels identify synthetic segments and do not establish observed sales or market behavior for the named medicines or locations.

**Table 4.** Descriptive properties of the synthetic panel.

| Property | Observed value |
| --- | --- |
| **Time span** | 1 January to 31 December 2023 |
| **Panel size** | 6 products × 5 regions × 365 days |
| **Records and fields** | 10,950 records and 21 fields |
| **Missing cells and duplicate date–product–region keys** | 0 and 0 |
| **Zero-sales observations** | 0 |
| **Daily units sold** | Mean 120.10; standard deviation 49.96; range 16 to 268 |

The panel contains temporal and commercial variation suitable for exercising lag features and calendar handling. Its completeness is also a limitation: it does not directly test missing feeds, intermittent demand, zero-sales periods, or late corrections. Only one annual cycle is represented. The generation procedure and random seed are not documented, so the observations can be characterized, but their generating process cannot be independently reconstructed. The descriptive patterns in this panel are not estimates of actual pharmaceutical demand.

### 9.2 Development progression and diagnostic status

Development explored a single boosted-tree predictor, a Prophet-assisted predictor, multiple search strategies, and weighted tree ensembles. Later experiments generated recursive 30-day forecasts and retained 14 rolling windows for each of 30 series. This history demonstrates that the forecasting task and output pathways were exercised in a synthetic setting. It does not demonstrate the full integration, review, and governance architecture proposed in this paper.

The historical experiment record contains a contemporaneous target contribution in a rolling feature, full-series fitting prior to some terminal assessments, and insufficient separation between model selection and later evaluation. Its residual-based forecast interval was not independently calibrated or tested for coverage. The affected numerical outputs are therefore excluded from claims of independent accuracy or model superiority.

The latter rolling summary contains 420 window-level diagnostics, with a macro mean RMSE of 12.55 units and macro MAPE and sMAPE of 10.09% and 9.71%, respectively. The mean series-level rolling RMSE ranges from 10.09 to 14.96 units. These are historical diagnostic quantities under the limitations above. They do not estimate the expected performance of a corrected pipeline or a future deployment.

### 9.3 What the present evidence supports

The evidence supports a descriptive synthetic panel, a product–region forecasting prototype, exploration of a statistical feature and tree ensembles, and retained forecast outputs. It does not establish controlled baseline superiority, incremental benefit from individual components, calibrated interval coverage, reproducibility of the synthetic generator, or benefit to planners. No real ERP deployment, prospective workflow evaluation, or financial outcome is reported here.

The reference protocol in Section 8 specifies how to obtain evidence for those stronger claims. It is intentionally presented as a protocol rather than retroactively attributed to the historical results. Corrected experiments require their own frozen data and result records before numerical claims can be updated.

## 10 Uncertainty and robustness

### 10.1 Calibration at the decision horizon

Intervals are associated with the target and forecast issuance procedure, not only with a model name. A residual distribution from one-step predictions cannot automatically calibrate a 30-day recursive path. Calibration should preserve the intended horizon and update policy, with horizon-specific or appropriately pooled residuals. Pooling across products or regions needs justification because sales scales and error distributions can differ substantially.

Time-series conformal methods offer useful approaches, but their validity depends on the specific method and its assumptions. For example, EnbPI studies the approximate validity of marginal coverage under specified temporal error conditions. Phorecaster365 does not claim that an arbitrary residual quantile or a generic train/calibration split inherits that result. A chosen interval method must be described accurately and tested under the dependence, refitting, and shift conditions of the intended deployment.

Marginal coverage across many forecast events differs from conditional coverage for a particular product and origin. Coverage on each day also differs from the probability that an entire 30-day path lies within the displayed bands. The interface must name the reported quantity. When the evidence is insufficient, an uncertainty warning and an approved fallback are more informative than an unsupported probability label.

### 10.2 Stress tests and expected system behavior

**Table 5.** Proposed robustness assessment.

| Stress condition | Assessment | Expected behavior |
|---|---|---|
| **Missing or delayed feed** | Remove observations or delay their knowledge time | Preserve missingness; hold affected scope or issue a labeled fallback |
| **Stockout periods** | Introduce availability constraints independently of latent demand | Identify censored sales; do not assert recovered demand without evidence |
| **Promotion or price change** | Alter timing, approval time, or magnitude | Enforce the information cutoff and separate scenarios |
| **New or discontinued product** | Shorten history or change lifecycle state | Apply a declared cold-start or run-out policy with review |
| **Returns and unit changes** | Add reversals or alter pack mapping | Preserve target semantics and reject unresolved unit conflicts |
| **Regional shift** | Change one segment while others remain stable | Detect localized degradation rather than relying on a portfolio average |
| **Model or export failure** | Timeout, incomplete horizon, malformed output, or duplicate retry | Prevent silent publication; preserve failure and acknowledgment state |

Synthetic stress tests are useful for verifying whether the intended response occurs. They cannot establish the frequency of those events in a real ERP environment. A future synthetic generator should document its distributions, dependencies, availability mechanism, event timing, seeds, and intentionally induced failure modes. In a censoring experiment, latent demand and realized sales should be stored separately to assess recovery accuracy. This is a proposed testing design and is not a reconstruction of the current generator.

## 11 Monitoring and governed operation

### 11.1 Monitoring with delayed outcomes

Data monitoring runs before outcome labels are available. It checks freshness, completeness, unit and schema changes, unexpected categories, and the population of series covered. Prediction monitoring can detect implausible values, abrupt changes, incomplete horizons, or widening intervals. Performance monitoring starts only after actuals mature under the stated finalization policy. A low error calculated against incomplete sales postings can create a misleading assurance of accuracy.

Drift indicators are investigation signals. A change in a feature distribution does not prove that predictive performance has deteriorated, and stable input distributions do not prove that the relationship between inputs and sales is unchanged. Reviews combine data changes, realized errors, interval behavior, and business events. Segment-level alerts preserve the location of a problem, while portfolio summaries quantify its extent.

Thresholds, review cadence, and escalation policies are set for the intended use and evaluated against false-alert burden. The report does not prescribe a universal accuracy target, drift threshold, or retraining frequency. These choices depend on historical variation, data latency, planning cadence, and the consequences of failure.

### 11.2 Retraining and release

Scheduled or event-triggered retraining creates a candidate, not an automatic replacement. The candidate uses a versioned snapshot and the declared evaluation procedure. The review compares it with the current model and fallback, checks for regressions in important segments and horizons, examines uncertainty behavior, and verifies operational compatibility. Promotion records the approving authority and release conditions.

Rollback preserves the prior model together with the compatible feature definition, calibration state, context schema, and hierarchy mapping. A model file alone is not sufficient to reproduce the earlier service. Forecasts already consumed downstream are not erased; a correction or superseding publication identifies the affected versions and required follow-up.

### 11.3 Governance and access

Governance is organized around explicit ownership of data meaning, model behavior, planning review, operations, and release approval. The NIST AI Risk Management Framework provides a general organizing reference through its govern, map, measure, and manage functions. Model cards document intended use, evaluation, restrictions, and monitoring, while dataset documentation records provenance, composition, transformations, and known gaps. Referencing these practices does not establish certification or compliance.

The audit history records source revisions, validation decisions, forecast generation, adjustments, approvals, publication acknowledgments, incidents, and model changes. Access is restricted by organization and role, with protection for commercially sensitive sales and pricing data. Future deployments should minimize the collection of unnecessary personal information and define retention, access review, and export controls. No patient-level data are required for the forecasting task described here.

## 12 Deployment pathway and prospective validation

The proposed pathway contains five stages, shown in Figure 4. Progression is based on evidence for the next use, rather than the existence of an attractive forecast plot or a favorable development metric. A failed assessment may return a component to an earlier stage, and a later operational incident may suspend publication while investigation proceeds.

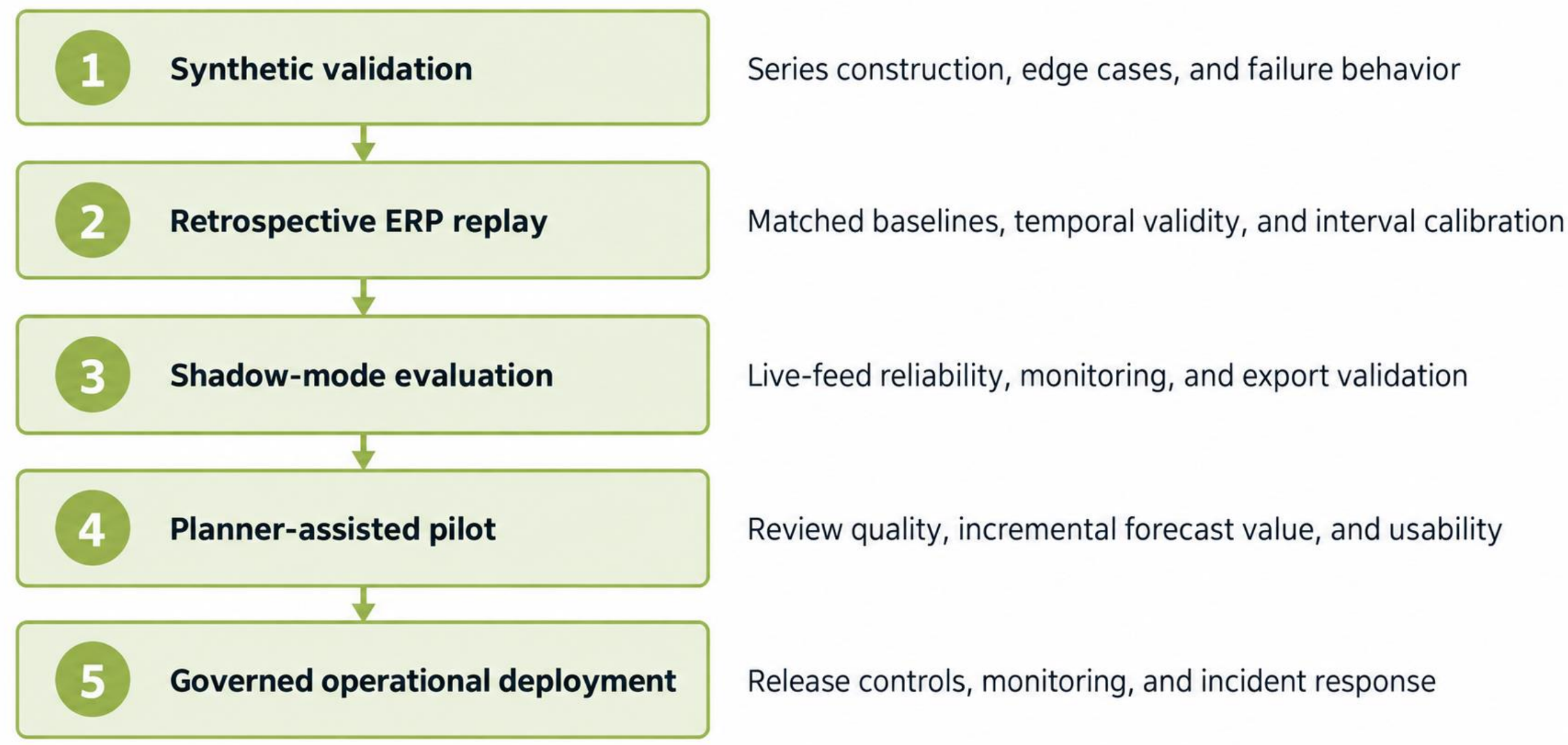

**Figure 4.** Evidence requirements across the deployment pathway. Each transition requires review of the stated evidence. The synthetic development record does not establish completion of the later stages.

Synthetic testing exercises data contracts, forecast generation, and failure behavior. Retrospective ERP replay then evaluates historical issuance using real operational semantics and established baselines. Shadow operation tests the live data and computation pathway while forecasts remain outside active planning decisions. A bounded planner-assisted pilot measures review behavior and decision-support usefulness. Governed operation requires ongoing monitoring, change control, and the ability to suspend or reverse a release.

The retrospective study should include multiple seasonal cycles where feasible, representative product and region coverage, stockout and return information, and historical versions of commercial plans. It should report exclusions and the fraction of operational series for which the forecast is supported. Performance on a clean subset cannot be generalized to excluded products without evidence.

The pilot should define its eligible users, products, decision horizon, review rules, and comparison process before operation. Relevant endpoints include review time, unresolved exceptions, forecast value added, data failures, and inappropriate reliance on unsupported outputs. A study of service levels, stockouts, expiry-related waste, or financial outcomes must additionally control for changes in ordering policy, supply availability, and other interventions. Forecast error improvements alone do not establish those benefits.

## 13 Discussion and limitations

Phorecaster365 treats temporal and operational provenance as part of the forecasting task. Its architecture connects a source snapshot to a reviewable output through a stable representation of the target, covariates, data states, model execution, and human decisions. This structure can support a meaningful comparison between algorithms because it holds the surrounding task definition and information boundary explicit.

The design also introduces costs. Versioned historical snapshots, approval records, and calibration evidence require storage and operational discipline. More granular monitoring creates additional review work. A strict temporal split can leave too little data for model fitting or uncertainty estimation. Local models are easy to associate with individual series but may be unstable with short histories; pooled models offer sharing while introducing new evaluation and isolation requirements. These trade-offs should be measured in the intended setting rather than resolved by architectural preference alone.

The present evidence is limited to a clean, one-year synthetic panel and historical development outputs. The generator is not documented; the clinical and commercial labels do not establish real-world representativeness, and the retained predictive metrics are affected by evaluation limitations. No independent calibration study, controlled ablation, user study, live ERP integration assessment, or prospective outcome evaluation is reported. The proposed evidence packages, review controls, and release pathway therefore remain design specifications requiring implementation and validation.

Source code, pseudocode, hyperparameter listings, and proprietary implementation details are not disclosed. The report provides sufficient conceptual detail to evaluate the architectural choices, but it does not provide an independently executable reproduction of the prototype. Public data and reproducible records of results would strengthen subsequent empirical work. A claim that the architecture is the first or uniquely capable solution would also require a broader comparative systems review and is not made here.

Future work should prioritize a frozen retrospective ERP study, calibrated multi-horizon forecasts, and a monitored planner-assisted pilot. Extensions to intermittent demand, pooled learning, direct forecasting, joint uncertainty, and hierarchical reconciliation can then be compared within the same evidence framework. More complex model families are justified only if they improve the relevant accuracy, reliability, or workflow outcomes under that evaluation.

## 14 Conclusion

Phorecaster365 is a human-supervised reference architecture for hybrid pharmaceutical sales forecasting and planning decision support. It defines how enterprise records become temporally valid forecast inputs, how model predictions retain their supporting evidence and uncertainty status, and how planner review leads to a versioned publication. The synthetic development record establishes the context for the design while leaving real-world accuracy and operational benefit unresolved. The proposed evaluation and deployment pathway provides concrete requirements for testing those outcomes before reliance on planning.

## Declarations

### Data availability

The development data described in this report are synthetic. No public data repository or generation procedure accompanies this version. The descriptive statistics and historical diagnostic summaries are reported in the manuscript; they do not constitute a reproducible data-generation package.

### Code availability

Source code and implementation artifacts are not publicly released with this report. The disclosure covers the reference architecture, conceptual methodology, evidence boundaries, and proposed validation process.

### Competing interests

The authors are affiliated with Izlek Technology and Xylemed. These commercial affiliations are relevant to the further development and potential commercialization of Phorecaster365 and should be considered when interpreting the report.

### Funding

No external grant funding was received for this work. The development of Phorecaster365 was undertaken by the authors as part of their professional activities at Izlek Technology and Xylemed.

### Author contributions

Houman Kazemzadeh: Conceptualization, methodology, software, formal analysis, validation, visualization, system architecture, and writing—original draft. Kamyar Naderi: Conceptualization, requirements definition, project administration, and writing—review and editing. Both authors reviewed and approved the final manuscript.

### Ethics and scope

The work described here uses synthetic aggregate sales records and does not report research involving human participants, identifiable patient records, or a clinical intervention. No formal ethics committee determination is claimed. Any subsequent study involving institutional data or participants requires the applicable institutional review and data governance processes.

## Appendix A: Synthetic Data Dictionary

The following groups account for all 21 fields in the development panel. They describe the data artifact; production use requires definitions, units, temporal provenance, and an approved purpose for each included field.

**Table A1.** *Fields in the synthetic ERP panel.*

| Field group | Fields | Interpretation and treatment |
| --- | --- | --- |
| **Series identity** | date; product; region | Daily product–region key; production identifiers and calendar require versioning |
| **Product and channel** | formulation; manufacturer; sales_channel; storage_condition | Product, commercial, or logistics context; effective dates are required |
| **Target and returns** | units_sold; units_returned | Preserve gross sales and return semantics; define any net-sales target separately |
| **Price and commercial plan** | price_usd; promo_flag | Price units and currency must be explicit; future promotion inputs require knowledge times |
| **Availability** | inventory_level | Potential availability context; does not quantify lost sales by itself |
| **Market and regulatory context** | prescription_type; insurance_coverage; prescriber_specialty; regulatory_status; prescription_origin | Synthetic categories; production definitions and admissibility need assessment |
| **Synthetic clinical context** | avg_patient_age; adherence_level; side_effect_flag; therapy_duration | Synthetic contextual fields; no patient-level or causal interpretation is supported |

### Appendix B: Historical diagnostic record

Table B1 preserves selected development summaries without ranking models by validated performance. Values are reproduced from the retained development record and are not recomputed under the proposed confirmatory protocol. MAE and RMSE are in units sold; comparisons are not controlled ablations.

**Table B1.** *Selected historical development diagnostics.*

| Development configuration | Mean MAE | Mean RMSE | Status |
|---|---|---|---|
| **XGBoost-only development reference** | 9.10 | 11.35 | Descriptive historical output |
| **XGBoost with statistical forecast signal** | 8.83 | 11.05 | Descriptive historical output |
| **Initial weighted tree ensemble** | 8.21 | 10.29 | Descriptive historical output |
| **Later ensemble variant 2** | 7.98 | 10.01 | Descriptive historical output |
| **Later ensemble variant 3** | 7.96 | 10.01 | Descriptive historical output |
| **Full-series terminal-fit variant** | 5.52 | 6.90 | In-sample assessment; excluded from accuracy claims |

***Note.*** Values are reproduced from the retained development record and were not recomputed under the proposed confirmatory protocol. MAE and RMSE are expressed in units sold. These comparisons are descriptive historical outputs and do not constitute controlled ablations or validated performance rankings.

The later rolling record comprises 420 rolling-origin windows across 30 series, with 14 windows per series. Its macro-average RMSE is 12.55 units, while the mean series-level rolling RMSE ranges from 10.09 to 14.96 units. These results are distinct from the terminal comparisons reported in Table B1 and remain affected by the historical feature-generation and model-selection limitations. The approximate 90% residual-derived interval used previously for visualization has not undergone independent coverage validation. Accordingly, no calibrated forecast interval, validated coverage claim, or controlled ablation result is inferred from these outputs.

### Appendix C: Minimum release evidence

A candidate release should identify the intended target and users, the approved data contract, temporal reconstruction evidence, the benchmark and ablation record, calibration and robustness results, and the supported product and region scope. It should also include the review workflow, publication and acknowledgment tests, incident and rollback procedures, and named ownership of residual limitations. The release decision records whether the candidate is approved, restricted, held, or rejected, together with its conditions and review date. This is a proposed acceptance framework; completion is not claimed for the present development system.